\documentclass[a4paper,twoside]{article}

\usepackage{epsfig} 
\usepackage{subcaption} 
\usepackage{calc} 
\usepackage{amssymb} 
\usepackage{amstext} 
\usepackage{amsmath} 
\usepackage{amsthm, amsfonts} 
\usepackage{multicol} 
\usepackage{pslatex} 
\usepackage{apalike} 
\usepackage[table,xcdraw]{xcolor} 
\usepackage{float} 
\usepackage{graphicx} 
\usepackage[space]{cite} 
\usepackage{multirow} 
\usepackage{booktabs} 
\usepackage[normalem]{ulem} 

\usepackage[pagebackref=true,breaklinks=true,colorlinks,bookmarks=false]{hyperref}

\usepackage{SCITEPRESS} 

\begin{document}

\title{A Comprehensive Study of Vision Transformers on Dense Prediction Tasks}

\author{\authorname{Kishaan Jeeveswaran, Senthilkumar Kathiresan, Arnav Varma, Omar Magdy,  Bahram Zonooz and Elahe Arani}
\affiliation{Advanced Research Lab, NavInfo Europe, Eindhoven, The Netherlands}
\email{\{firstname.lastname\}@navinfo.eu, bahram.zonooz@gmail.com}
}


\abstract{
Convolutional Neural Networks (CNNs), architectures consisting of convolutional layers, have been the standard choice in vision tasks. 
Recent studies have shown that Vision Transformers (VTs), architectures based on self-attention modules, achieve comparable performance in challenging tasks such as object detection and semantic segmentation. 
However, the image processing mechanism of VTs is different from that of conventional CNNs.
This poses several questions about their generalizability, robustness, reliability, and texture bias when used to extract features for complex tasks. 
To address these questions, we study and compare VT and CNN architectures as a feature extractor in object detection and semantic segmentation. 
Our extensive empirical results show that the features generated by VTs are more robust to distribution shifts, natural corruptions, and adversarial attacks in both tasks, whereas CNNs perform better at higher image resolutions in object detection.
Furthermore, our results demonstrate that VTs in dense prediction tasks produce more reliable and less texture biased predictions.
}

\onecolumn \maketitle \normalsize \setcounter{footnote}{0} \vfill

\begin{figure*}[tb!h]
\centering
   \includegraphics[width=.9\linewidth]{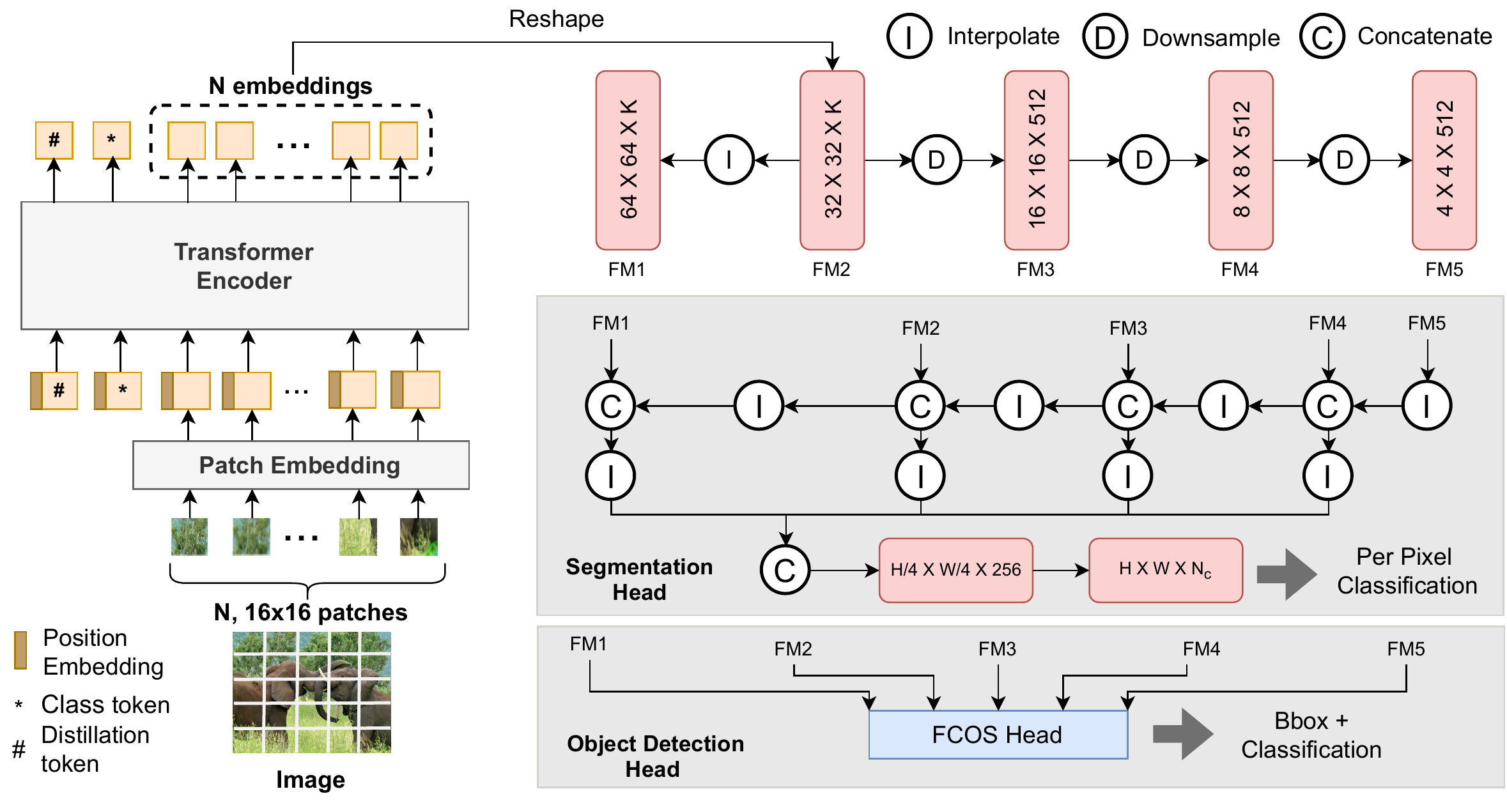}
   \caption{Architecture for object detection and semantic segmentation with VT backbone.}
\label{figure:architecture}
\end{figure*}


\section{\uppercase{Introduction}}
\label{sec:introduction}

CNNs' remarkable performance has made them the prominent choice of architecture in computer vision tasks \cite{ResNet, efficientnet}.
On the other hand, the Transformers have become dominant in NLP owing to their ability to learn long-term dependencies via self-attention.
Recent breakthrough of VTs \cite{dosovitskiy2020image} demonstrated that the Transformer-based architecture can also be applied to image classification.
This motivated the vision community to adapt the self-attention based architectures as feature extractors for more complex tasks such as depth prediction \cite{ranftl2021vision}, object detection, and semantic segmentation \cite{liu2021swin, srinivas2021bottleneck}.
VTs have achieved compelling performance in these tasks, presenting them as an alternative architectural paradigm.

However, the mechanism by which VTs process the images is significantly different from that of well-studied CNNs.
CNNs use a sequence of convolutional layers to extract features with progressively increasing receptive field.
These convolutional layers have inherent inductive biases, such as locality and translation equivariance, that are helpful for computer vision tasks. 
However, their local receptive field makes them incapable of capturing the global context.
VTs, on the other hand, split the input image into non-overlapping tokens, and use a sequence of self-attention modules to process these tokens.
These self-attention modules have global receptive field, but they lack the inductive biases inherent in convolutional layers, making them data hungry. 
Therefore, the choice of these two architectures comes with their own merits and limitations.

The difference in the fundamental working principles of VTs compared to CNNs raises many questions: How well do they perform for distribution shifts? How robust are they to the adversarial attacks? How reliable are their predictions for real-world applications? To what extend they learn the shortcuts, such as the texture of an object, rather than the intended solution?
Although few of these questions are addressed in the image classification domain \cite{bhojanapalli2021understanding, paul2021vision}, they have not yet been fully addressed for complex tasks including detection and segmentation.
To this end, we perform an in-depth analysis by constructing simple detection and segmentation models using DeiT \cite{touvron2020training} as the feature extractor for dense prediction tasks. 
Our contributions are as follows:
\begin{itemize}
    \item Evaluating on in-distribution-dataset, we find VTs are more accurate but slower than CNNs counterparts. In addition, the results on out-of-distribution (OOD) dataset reveals that VTs are also more generalizable to distribution shift.
    \item Our results show that VTs are better calibrated and thus, more reliable than CNNs, thereby making them better suited for deployment in safety-critical applications.
    \item Increasing the inference image resolution, we show that the performance of both VTs and CNNs degrade. However, in detection, CNNs outperform their VT counterparts at higher resolutions.
    \item We demonstrate that VTs converge to wider minima compared to CNNs, which we attribute to their generalizability.
    \item We show that VTs are consistently more robust to natural corruptions and (un)targeted adversarial attacks than CNNs.
    \item We extend the texture bias study \cite{bhojanapalli2021understanding} for dense prediction tasks. The results show that the VTs are less dependent on texture cues than CNNs to make their predictions.
\end{itemize}

\section{\uppercase{Related Work}}
Transformer architectures adapted for image classification, such as ViT~\cite{dosovitskiy2020image} and DeiT~\cite{touvron2020training}, have achieved comparable performance to state-of-the-art CNNs. Later methods modified these vision transformers(VTs) to act as feature extractors in dense prediction tasks such as object detection, semantic segmentation, and depth prediction~\cite{wang2021pyramid,liu2021swin,ranftl2021vision}. The progress of VTs present them as an alternative architecture to CNNs across vision tasks.

To study the impact of architecture change, recent works have compared VTs and CNNs on aspects beyond speed and accuracy for \textit{image classification}. 
Among these works, studies such as ~\cite{bhojanapalli2021understanding} and \cite{paul2021vision} have compared VTs and CNNs in terms of robustness to adversarial attacks and natural corruptions.
\cite{naseer2021intriguing} further studied the texture-bias of VTs and CNNs for image classification. \cite{minderer2021revisiting} additionally investigated the model calibration of VTs and CNNs, and demonstrated that type of architecture is a major determinant of properties of calibration.
However, there has been no study of the impact of VTs on generalizability, robustness, calibration, and texture-bias in \textit{dense prediction tasks}, including detection and segmentation, when replacing CNNs as the feature extractor.

We perform an exhaustive comparison of VTs and CNNs for their generalizability to higher resolutions and distribution shifts, robustness to adversarial attacks and natural corruptions, reliability, and texture-bias for object detection and semantic segmentation.

\section{\uppercase{Methodology}}
We conduct the comprehensive empirical study on object detection and semantic segmentation tasks with CNN and VT backbones of different sizes. Here, we explain the details of architecture used in this study.

\begin{table*}[!htbp]
\centering
\caption{Comparison of VTs and their CNN counterparts for object detection on COCO dataset and semantic segmentation on COCO-Stuff dataset. Best score for each metric is in bold.}
\label{tab:coco-det-seg}
\resizebox{\linewidth}{!}{
\begin{tabular}{l|ccccc||ccccc}
\hline
\multirow{3}{*}{Backbone} & \multicolumn{5}{c||}{Detection} & \multicolumn{5}{c}{Segmentation} \\ \cline{2-11}
 & \begin{tabular}[c]{@{}c@{}}\#Param\\ (M)\end{tabular} & \begin{tabular}[c]{@{}c@{}}MAC\\ (G)\end{tabular} & \begin{tabular}[c]{@{}c@{}}Inf. \\ Time(ms)\end{tabular} & mAP & \begin{tabular}[c]{@{}c@{}}Energy \\ (kJ) \end{tabular} & \begin{tabular}[c]{@{}c@{}}\#Param\\ (M)\end{tabular} & \begin{tabular}[c]{@{}c@{}}MAC\\ (G)\end{tabular} & \begin{tabular}[c]{@{}c@{}}Inf. \\ Time(ms)\end{tabular} & mIoU & \begin{tabular}[c]{@{}c@{}}Energy \\ (kJ) \end{tabular} \\ \hline
RN-18 & 36 & 122.34 & \textbf{32.29} & 26.04 & \textbf{4.6} & 15 & 17.76 & \textbf{6.64} & 30.28 & \textbf{1.7} \\
DeiT-T & \textbf{30} & \textbf{107.06} & 41.09 & 38.19 & 5.7 & \textbf{11} & \textbf{2.51} & 11.57 & 35.08 & 2.1 \\\hline
RN-50 & 49 & 141.00 & 38.95 & 39.45 & 5.3 & 28 & 35.02 & 12.83 & 35.18 & 2.4 \\
DeiT-S & 49 & 109.58 & 46.85 & 42.56 & 6.2 & 28 & 4.59 & 18.12 & 39.79 & 2.7 \\\hline
RNX-101 & 114 & 193.54 & 60.92 & 41.20 & 8.3 & 93 & 87.84 & 27.04 & 38.02 & 5.2 \\ 
DeiT-B & 120 & 116.74 & 72.33 & \textbf{45.91} & 9.8 & 100 & 12.38 & 38.14 & \textbf{41.20} & 6.6 \\
\hline
\end{tabular}}
\end{table*}

\subsection{Transformer as a Feature Extractor}
We use Data Efficient Image Transformer (DeiT) \cite{touvron2020training} as our VT feature extractor. In DeiT, the input image is divided into $N$ non-overlapping patches of a fixed size ($16 \times 16$) and the patches are flattened and embedded using a linear layer to $K$ dimensions. 
A position embedding is added element-wise to the patch embeddings to help the model to understand some notion of the order of the input patches. The resulting tensor is given as input to repeated blocks of self-attention and feedforward layers. 
The final representation of class encoding is passed through a feedforward network (FFN) before feeding it to a softmax layer to infer the classes. 
The class encoding learns the context and class specific information from the image patches. 

We modify DeiT to make it suitable for object detection and semantic segmentation by removing the \textit{FFN} and refining the outputs of the final block ($N + 2$ embeddings) before passing on to the \textit{heads}.
The refinement process includes removing the class and distillation embeddings (used in DeiT for classification) after which N embeddings remain. 
These N embeddings are reshaped into a feature map of $32 \times 32 \times K$ (FM2 in Figure \ref{figure:architecture}). A series of convolutional downsampling layers are used to create multiple feature maps of spatial dimensions 16 (FM3), 8 (FM4), and 4 (FM5) from FM2. 
Finally, FM2 is upsampled to obtain a feature map of spatial dimension 64 (FM1).
These five feature maps are then passed to the prediction head of the models.

\subsection{Detection Head}
For object detection, we use Fully Convolutional One-Stage object detector (FCOS) \cite{Tian2019}, an anchor-free method that makes predictions based on key-point estimation, and is one of the state-of-the-art methods.
The detection head infers the classification score, bounding box parameters, and centerness score, and is shared between feature maps at multiple scales (FM1 to FM5) as shown in Figure \ref{figure:architecture}. Since it is a pixel-wise dense bounding box predictor, the centerness score is used to suppress low quality bounding boxes which are predicted at pixel locations far away from the object center.

\subsection{Segmentation Head}
For segmentation, the outputs of the backbone are passed to a light-weight segmentation head to infer dense pixel-wise classification scores.
In the segmentation head, starting from the smallest spatial resolution, every feature map is interpolated and concatenated in the channel dimension with the adjacent larger feature map. The same approach is adopted for the subsequent feature maps. These feature maps are bilinearly interpolated to one-fourth of the input resolution, and concatenated. Finally, bilinear interpolation is used to upsample the resultant feature map to the input resolution, which predicts the class probabilities for every pixel. 
\section{\uppercase{Experimental Setup}}
The experiments are conducted on both detection and segmentation tasks for VT and CNN backbones of different network sizes. We use three DeiT variants as VT backbone - Tiny(T), Small(S), and Base(B) with input patch size 16$\times$16 - and three CNN counterparts with the same range of parameters - ResNet-18 (RN-18), ResNet-50 (RN-50), ResNeXt-101 [32$\times$8d] (RNX-101).

\noindent
\textbf{Training Dataset.} 
The detection models are trained and evaluated on the COCO dataset \cite{lin2014microsoft} which consists of 81 classes. The segmentation models are trained and evaluated on COCO-Stuff \cite{caesar2018coco} dataset which contains $172$ classes - $80$ "things" classes, $91$ "stuff" classes, and 1 unlabelled class. The datasets consist of 118K training images and 5K validation images. 
We choose COCO dataset for our experiments because it is a challenging benchmark dataset with common and naturally occurring real-world scenes, making it suitable for comparative experiments on dense prediction models.

\noindent
\textbf{Training Details. } 
All models are trained on a Tesla V100 GPU at $512\times512$ resolution using AdamW optimizer \cite{loshchilov2019decoupled} with an initial learning rate of $5e^{-4}$, weight decay of $0.05$, and a cosine learning rate scheduler. The networks with different backbones are trained with different batch sizes: DeiT-B and RNX-101 with 8, DeiT-S and RN-50 with 16, and DeiT-T and RN-18 with 32. The detection models are trained for 55 epochs and the segmentation models are trained for 45 epochs.
The data augmentation includes random horizontal flip, random crop, and random photometric distortions such as random contrast $\in [0.5, 1.5]$, saturation $\in [0.5, 1.5]$ and hue $\in [-18, +18]$. 
We use Imagenet~\cite{deng2009imagenet} pretrained weights for initializing all the backbones.

\noindent
\textbf{Evaluation Metrics. } 
The metrics used to measure the performance of segmentation (SEG) and detection (DET) models are mIoU (mean Intersection over Union) and mAP (mean Average Precision @0.5:0.95 IoU), respectively, unless stated otherwise.
In addition to these accuracy metrics, we report number of learnable parameters (in millions), Multiply-Accumulate operations (GMAC) for the architecture, inference time per image in milliseconds (ms), and inference energy consumption of a model (in kilo Joules).
We report the average inference time and total inference energy over 500 samples.
All metrics are calculated at the training resolution.

\section{\uppercase{Generalization}}
In this section, we probe VTs and CNNs for generalizability to in-distribution and OOD data. We also investigate the effect of input resolution on generalization.

\subsection{In-Distribution Evaluation}
As shown in Table \ref{tab:coco-det-seg}, the VT-based object detectors outperform their CNN counterparts, but at the cost of inference speed.

Now, MAC represents the computational complexity of the model, and is usually correlated with the inference speed.
Table~\ref{tab:coco-det-seg} shows that although VTs have less complexity than CNNs, they are slower than CNNs. 
This might be mainly due to the fact that GPUs are less optimized for the Transformers \cite{Ivanov2020} than CNNs. This could also explain the higher energy consumption of VTs.
Additionally, we note that the complexity of the largest VTs is less than that of the smallest CNN ($116$ vs $122$ GMAC).
Furthermore, the computational complexity of VTs does not increase as much as that of CNNs with number of parameters.
Similar to the results for object detection, the VT-based segmentation models outperform their CNN counterparts at the cost of inference speed and energy consumption. 

To summarize, VT-based models are more accurate than their CNN counterparts, but the CNN-based models are faster and consume less energy. 
However, since the VT-based models are less complex, we contend that they can be faster than their CNN counterparts if the GPUs are optimized for the VT architectures \cite{Ivanov2020}.

\begin{figure}[t]
    \centering
    \includegraphics[width=1.0\linewidth]{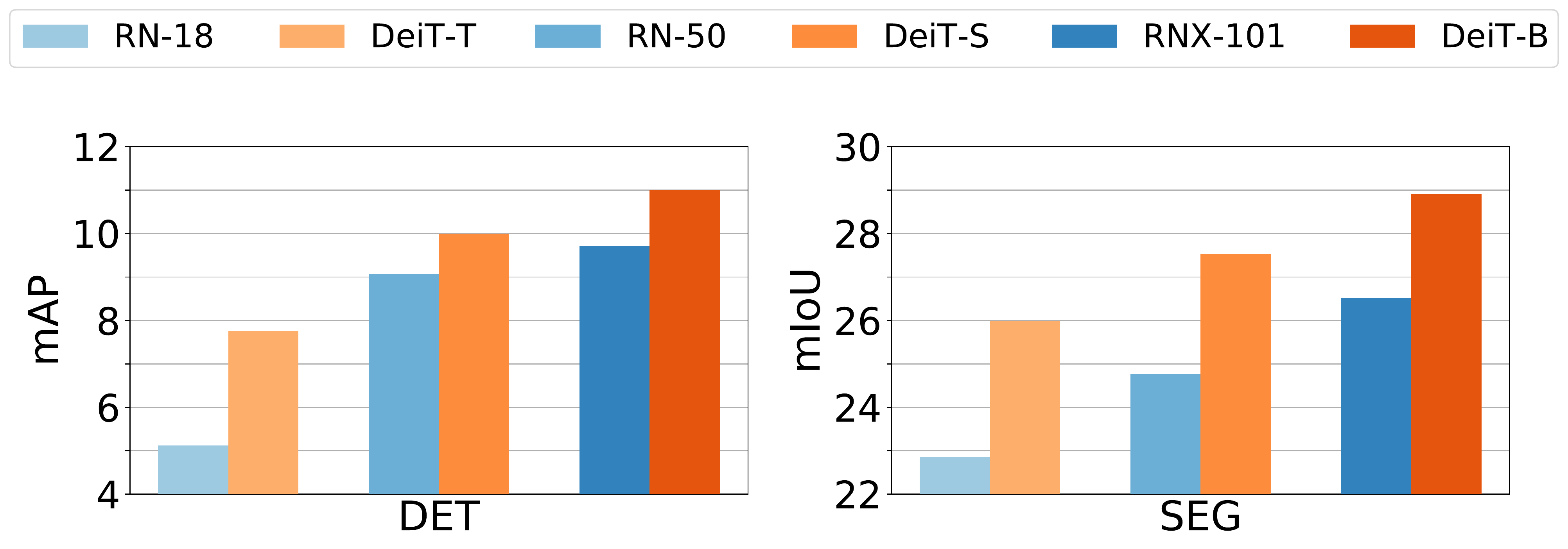}%
    \caption{Comparison of VTs and their CNN counterparts for  OOD performance. Object detection models are trained on COCO and evaluated on BDD100K for 8 classes. Semantic segmentation models are trained on COCO-Stuff and evaluated on BDD10K for 14 classes.}
    \label{fig:ood_fig}
\end{figure}

\begin{figure*}[t]
    \centering
    \begin{subfigure}[b]{\textwidth}
    \centering
    \includegraphics[width=0.5\linewidth]{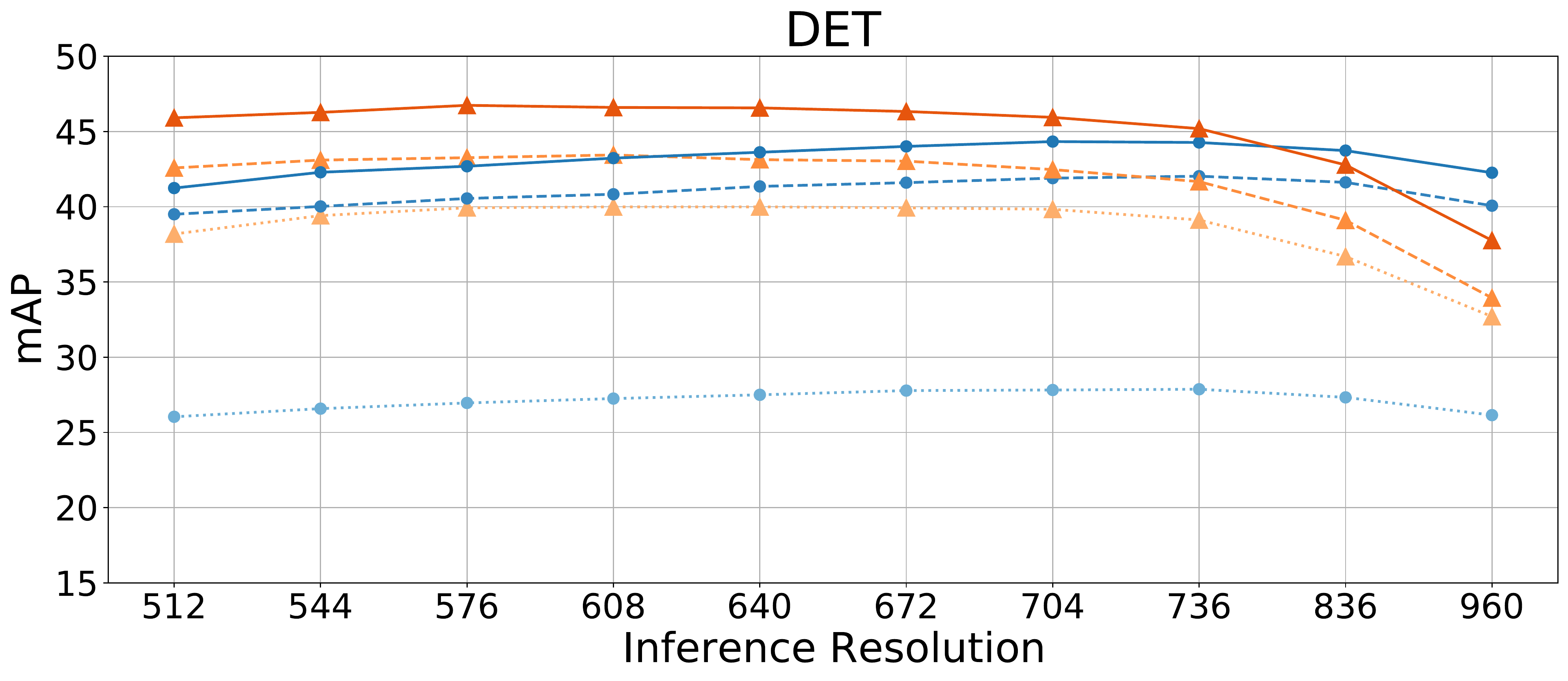}%
    \hfill
    \includegraphics[width=0.5\linewidth]{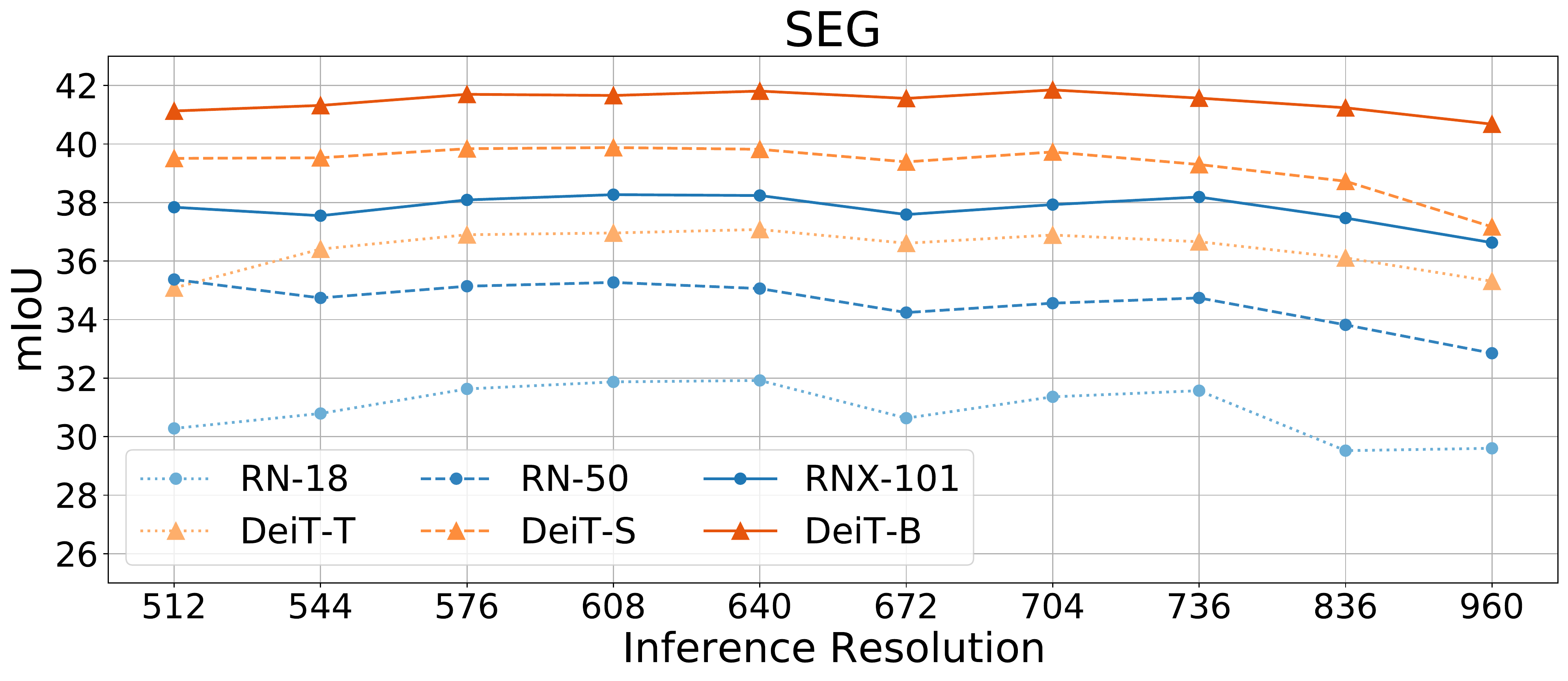}
    \end{subfigure}
    \caption{Comparison of VTs and their CNN counterparts at inference resolutions higher than training resolution ($512\times 512$).}
    \label{fig:inference_resolution_seg}
\end{figure*}

\subsection{Out-of-Distribution Evaluation}
\label{sec:ood}

Despite the good performance of the models on in-distribution data, it is important to evaluate how well they perform on unseen data, especially when they are deployed for real-world applications.
The performance of the model on such unseen data indicates its generalizability to OOD datasets.

The detection and segmentation models trained on COCO and COCO-Stuff are evaluated on BDD100K \cite{yu2020bdd100k} and BDD10K datasets, respectively.  
BDD dataset has a different distribution from that of COCO since it is composed of road scenes with traffic elements like pedestrians, vehicles, road, and traffic signs.
BDD100K for object detection has 10K test images consisting of $10$ classes, out of which, \emph{'rider'} and \emph{'traffic sign'} do not have a corresponding class in COCO.
So, we evaluate the models for $8$ matching classes on BDD100K.
BDD10K for semantic segmentation, is a subset of BDD100K, with 1k test images consisting of $19$ classes.
\emph{'pole'}, \emph{'traffic-sign'}, \emph{'vegetation'}, \emph{'terrain'}, and \emph{'rider'} classes do not have a corresponding class in COCO-Stuff. 
Thus, we evaluate for $14$ matching classes and map the COCO-Stuff classes which do not have a corresponding class in BDD10K to \emph{'unlabeled'} class.

Figure \ref{fig:ood_fig} shows that VTs achieve higher performance than their CNN counterparts in both tasks. 
In segmentation, DeiT-T and DeiT-S also outperform significantly larger CNN backbones (RN-50 and RNX-101, respectively). Similarly, in detection, DeiT-S outperforms RNX-101.
These results suggest that the features learned by the VT backbones are more generalizable to OOD data. We conduct further experiments in Section~\ref{sec:lanscape} to analyze the generalizability of these models.

\subsection{Inference Resolution Study}
Given the global receptive field of VTs~\cite{dosovitskiy2020image}, they should be able to handle larger inference resolutions better than CNNs.
Although this has been tested for depth estimation~\cite{ranftl2021vision}, it hasn't been tested for object detection and semantic segmentation.
Therefore, we compare the detection and segmentation performance of VT and CNN backbones when inferred at resolutions higher than the training resolution (512$\times$512).

When inferring at higher resolutions, the patch size of VTs is fixed at 16$\times$16 resulting in a larger sequence length. 
Though Transformer architectures can handle arbitrary sequence lengths, VTs need interpolation of the position embeddings to adapt to the new sequence length. We perform bicubic interpolation over the pretrained position embeddings~\cite{touvron2020training}.
CNNs, on the other hand, can infer at higher resolutions without any modifications.

Figure \ref{fig:inference_resolution_seg} shows that, in detection, the performance degradation at higher resolutions is more gradual for CNNs as compared to VTs. Consequently, CNNs outperform their VT counterparts at higher inference resolutions in detection.
However, this trend is not observed in semantic segmentation, where higher inference resolution has similar effect on both CNNs and VTs, and VTs outperform CNNs at all resolutions.
We believe that this is because the interpolated positional embeddings might not be as effective for detection as they are for segmentation.

Contrary to the conjecture made by \cite{ranftl2021vision} for depth estimation, the global receptive field of VTs does not provide an advantage over CNNs at higher inference resolutions for detection.
This difference in behaviour of VTs across tasks raises questions about the cross-task suitability of interpolating the position embeddings. We leave this analysis for future work.

\subsection{Convergence to Flatter Minima Analysis}
\label{sec:lanscape}

\begin{figure*}[htb]
    \centering
    \begin{subfigure}[b]{\textwidth}
    \centering
    \includegraphics[width=0.5\linewidth]{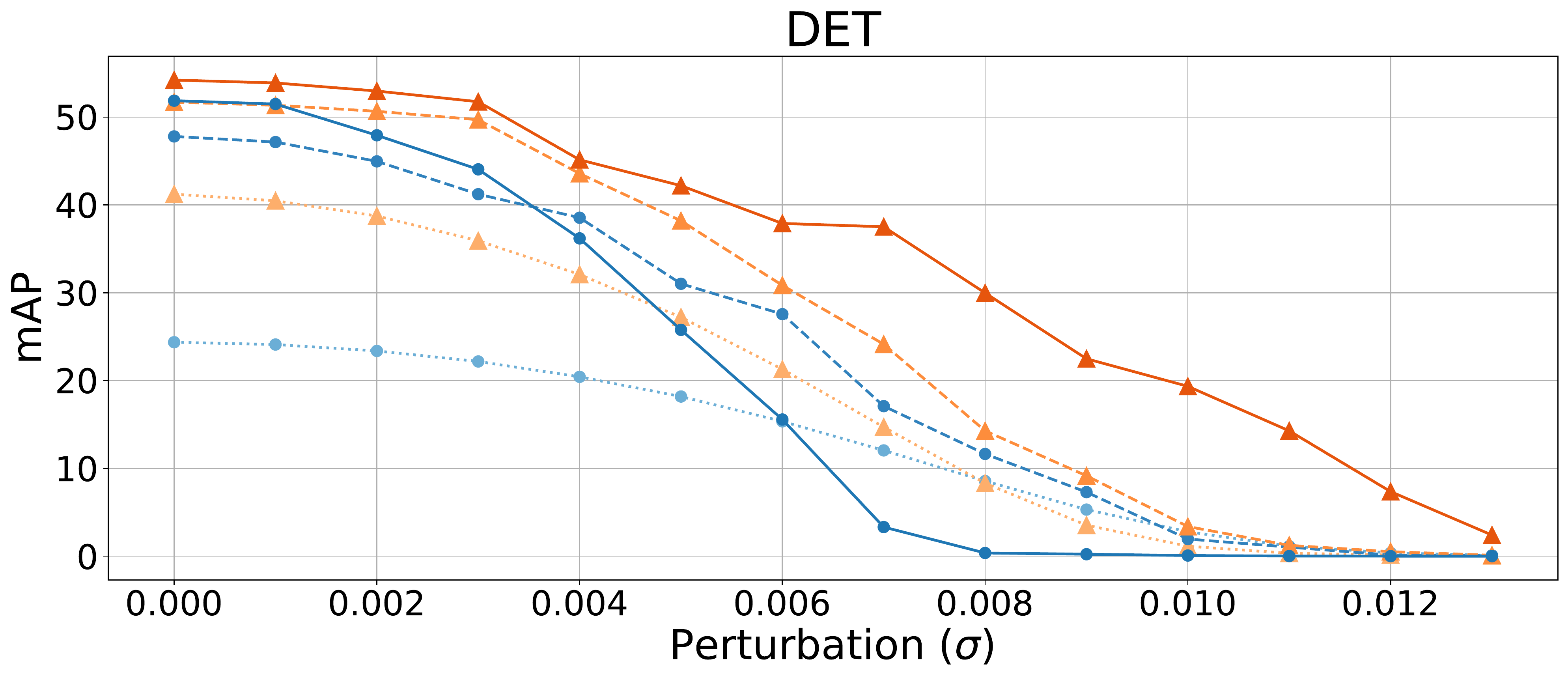}%
    \hfill
    \includegraphics[width=0.5\linewidth]{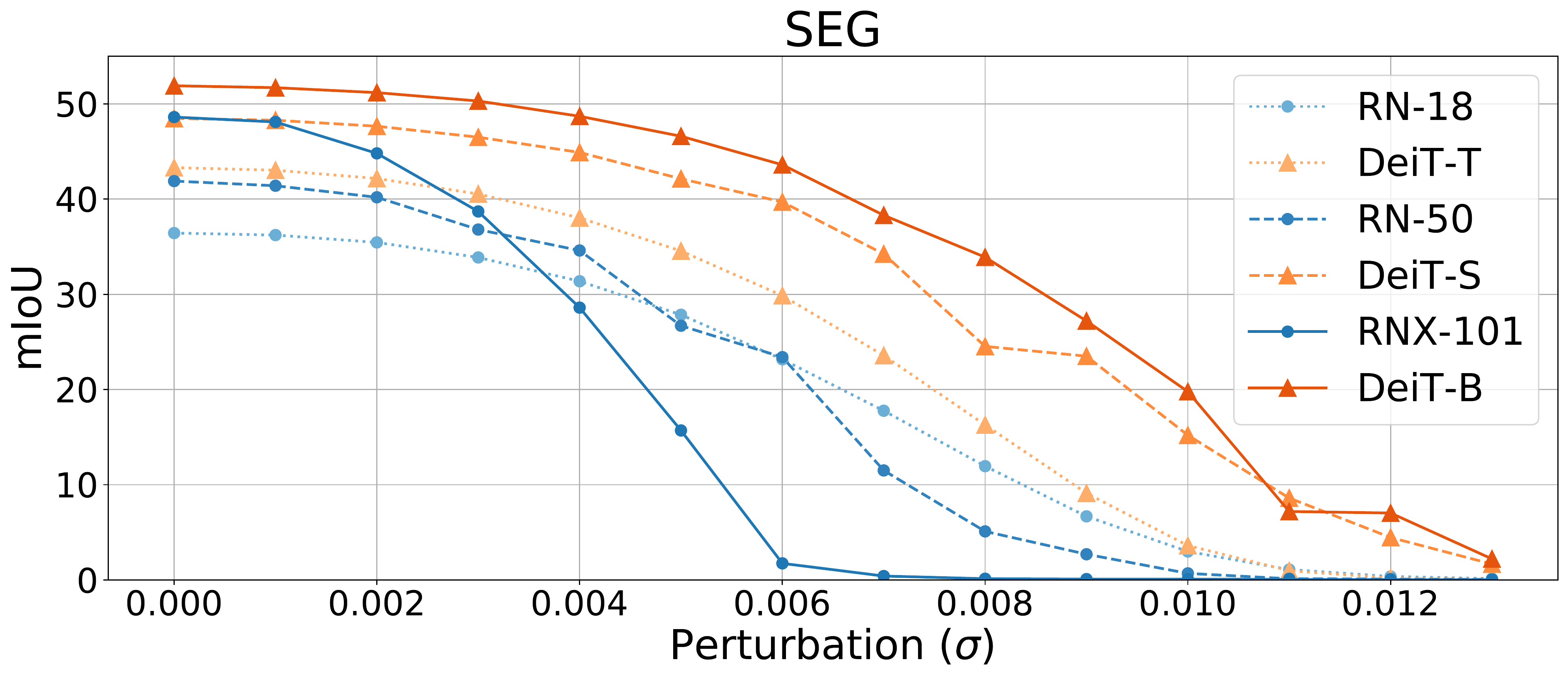}
    \end{subfigure}
    \caption{Comparison of train set performance of VTs and their CNN counterparts as a function of Gaussian noise added to the model parameters.}
    \label{fig:gaussian_noise}
\end{figure*}

\begin{figure*}[tb]
    \centering
    \begin{subfigure}[b]{\textwidth}
        \centering
        \includegraphics[width=0.5\linewidth]{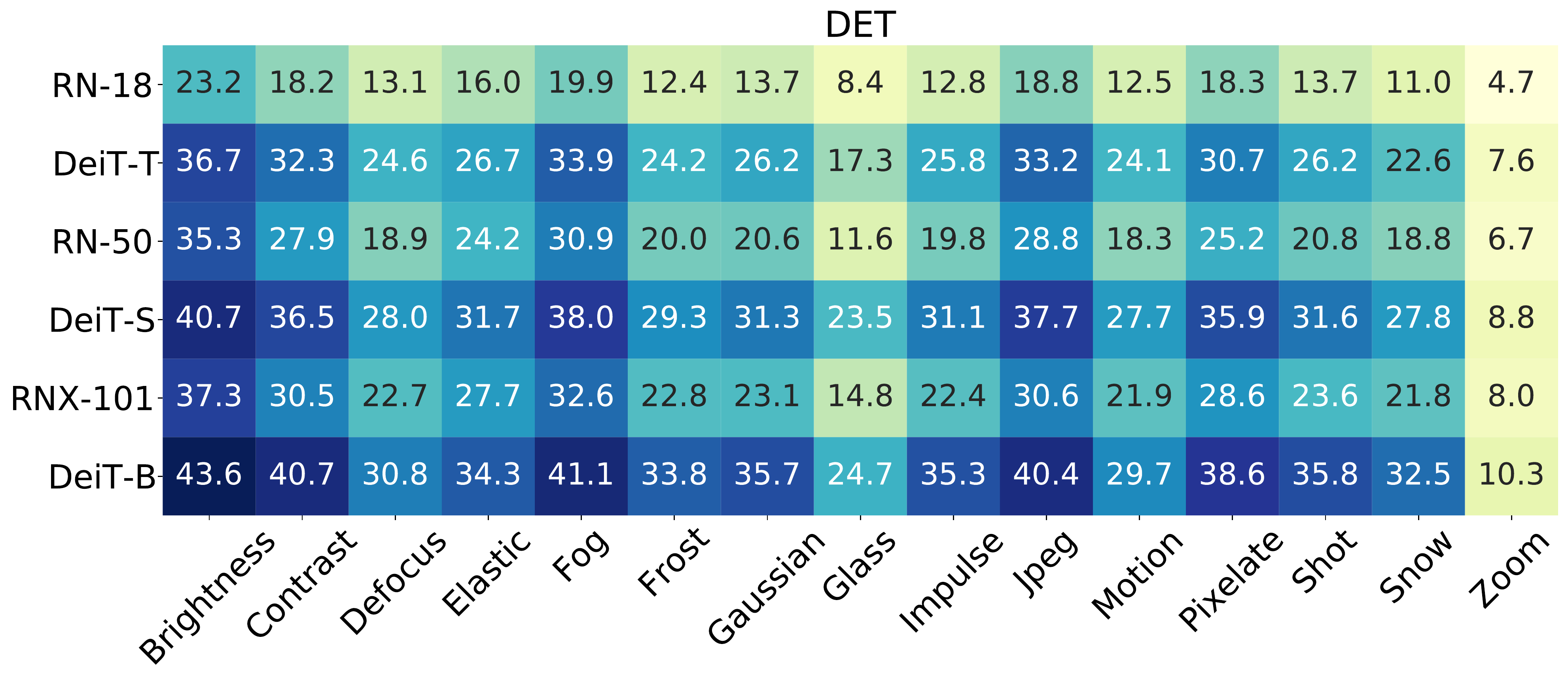}
        \hfill
        \includegraphics[width=0.48\linewidth]{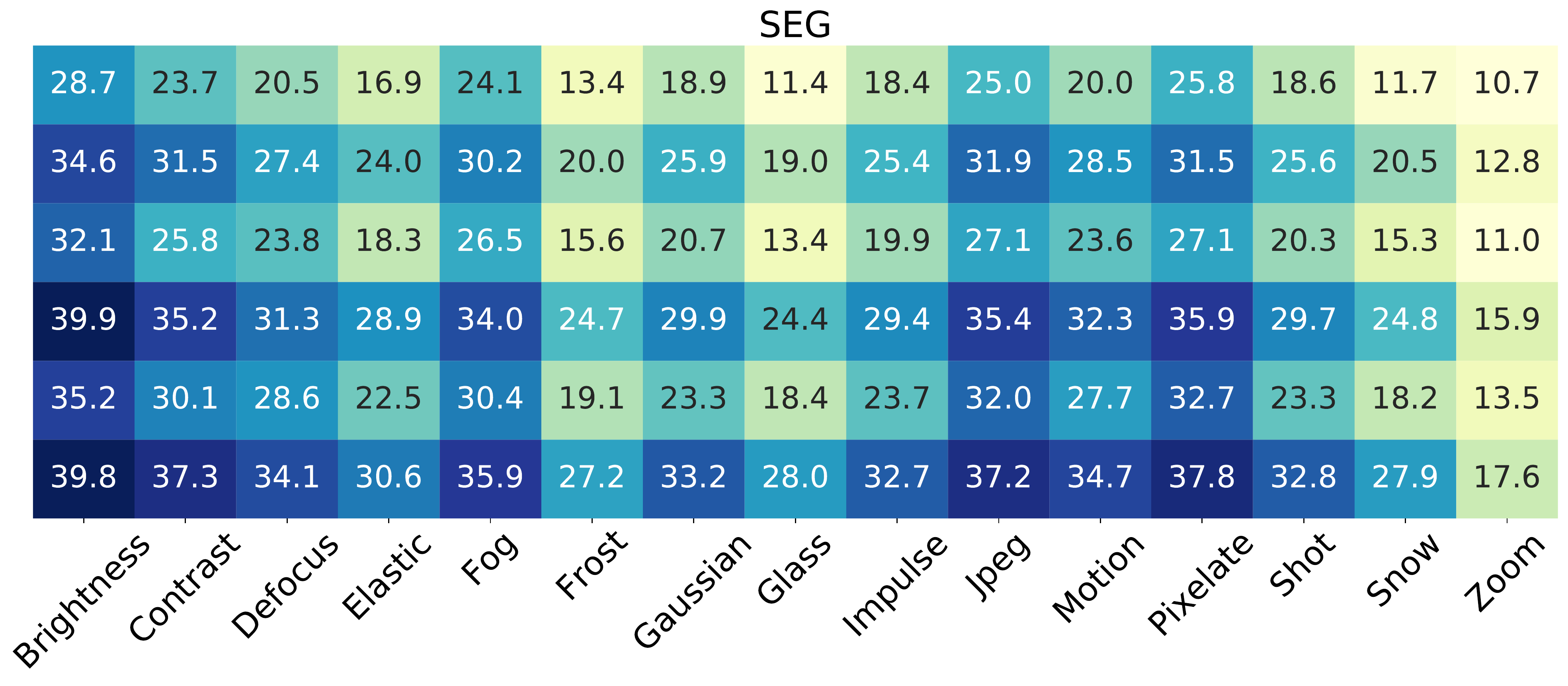}
    \end{subfigure}
    \caption{Performance comparison of VTs and their CNN counterparts under natural corruptions.}
    \label{fig:corrupt}
\end{figure*}

Since there are multiple solutions to the optimization objective of a model, the local geometry at the convergence point may affect the model's generalization. It has been shown that the models that converge to flatter minima in the loss landscapes are more robust to distribution shift, and hence more generalizable~\cite{keskar2016large, chaudhari2019entropy}. 
If models find solutions in flatter minima, the performance would not change significantly when the weights are perturbed. 
Meanwhile, if the models converge to sharper minima, even a slight perturbation in the weights could result in drastic changes in performance.

To analyze the generalizability of the trained models, we add noise with increasing strengths to their trained weights. The noise is sampled from a Gaussian distribution with mean 0 and standard deviation ranging from 0.0 to 0.013 in steps of 0.001. Finally, we infer these models with perturbed weights on $20\%$ of the training data.
As shown in Figure \ref{fig:gaussian_noise}, although performance of all detection and segmentation models degrades as noise increases, VTs perform better than CNNs.
It is interesting to note that in detection, DeiT-S and Deit-T have a sharper decline in performance compared to DeiT-B, whereas in segmentation, all three VTs have a similar decline.
Moreover, in both tasks, the performance of CNNs degrades much more sharply than that of the VTs, with the largest CNN backbone (RNX-101) showing the sharpest drop.
Hence, unlike VTs, an increase in CNN model-size doesn't necessarily lead to convergence to flatter minima.
The results demonstrate that VTs converge to flatter minima which could explain their ability to generalize better to unseen data compared to CNNs as seen in Section \ref{sec:ood}.

\section{Model Calibration}

The neural networks that are being used in safety-critical applications such as autonomous driving are expected to be accurate and reliable. Reliable models are well-calibrated, which means their prediction confidences and the accuracy of those predictions are highly correlated.
However, recent studies in classification have shown that highly accurate CNNs are poorly calibrated \cite{guo2017calibration}, and VTs are better calibrated than CNNs \cite{minderer2021revisiting}.
Here, we extend the reliability study of CNNs and VTs \cite{minderer2021revisiting} for detection and segmentation and report the results for in-distribution data.

Expected Calibration Error (ECE) and Maximum Calibration Error (MCE) \cite{naeini2015obtaining} are common metrics used to measure the calibration error of a neural network in classification. 
ECE is computed by binning the predictions based on the confidence score and taking the weighted mean of the difference between the average accuracy and confidence of each bin.
MCE, on the other hand, is the maximum difference between the average accuracy and confidence across all bins.
We use ECE and MCE with 15 bins to measure the miscalibration in segmentation models.
For object detection, Detection-ECE (D-ECE) and (w)D-ECE \cite{kuppers2020multivariate} are used to measure the calibration error. 
D-ECE extends ECE by including the bounding box information such as coordinates and scale of the bounding box as additional binning dimensions. 
(w)D-ECE takes the weighted average of D-ECE scores with respect to samples in each class. We use 15 bins, confidence threshold 0.3, and IoU threshold 0.6 in our analysis.

\begin{table*}[tb]
    \caption{Reliability comparison of VTs and their CNN counterparts. Best score for each metric is in bold.}
    \label{tab:calibration} \centering
    \begin{tabular}{llcccccc}
    \toprule
    Task & Metric & RN-18 & DeiT-T & RN-50 & DeiT-S & RNX-101 & DeiT-B \\
     \midrule
    \multirow{2}{*}{{DET}}  
    & (w)D-ECE & 0.238 & 0.193 & 0.200 & 0.193 & 0.219 & \textbf{0.165} \\
    & D-ECE  & 0.164 & 0.120 & 0.145 & 0.119 & 0.168 & \textbf{0.094} \\ \hline
    \multirow{2}{*}{{SEG}}
    & ECE & 0.157 & 0.153 & 0.159 & 0.158 & 0.163 & \textbf{0.147} \\
    & MCE & 0.378 & 0.371 & 0.389 & 0.378 & 0.397 & \textbf{0.369} \\
    \bottomrule
    \end{tabular}
\end{table*}

From the results in Table \ref{tab:calibration}, VTs are better calibrated than their CNN counterparts in both the tasks. However, there is no relationship observed between model-size and calibration within either VTs or CNNs.
Hence, in detection and segmentation, the calibration of a model is mainly determined by its architecture, and not by its size. These observations are in line with the results in image classification~\cite{minderer2021revisiting}.

\begin{figure*}[htb]
    \centering
    \begin{subfigure}[b]{\textwidth}
        \centering
        \includegraphics[width=0.5\linewidth]{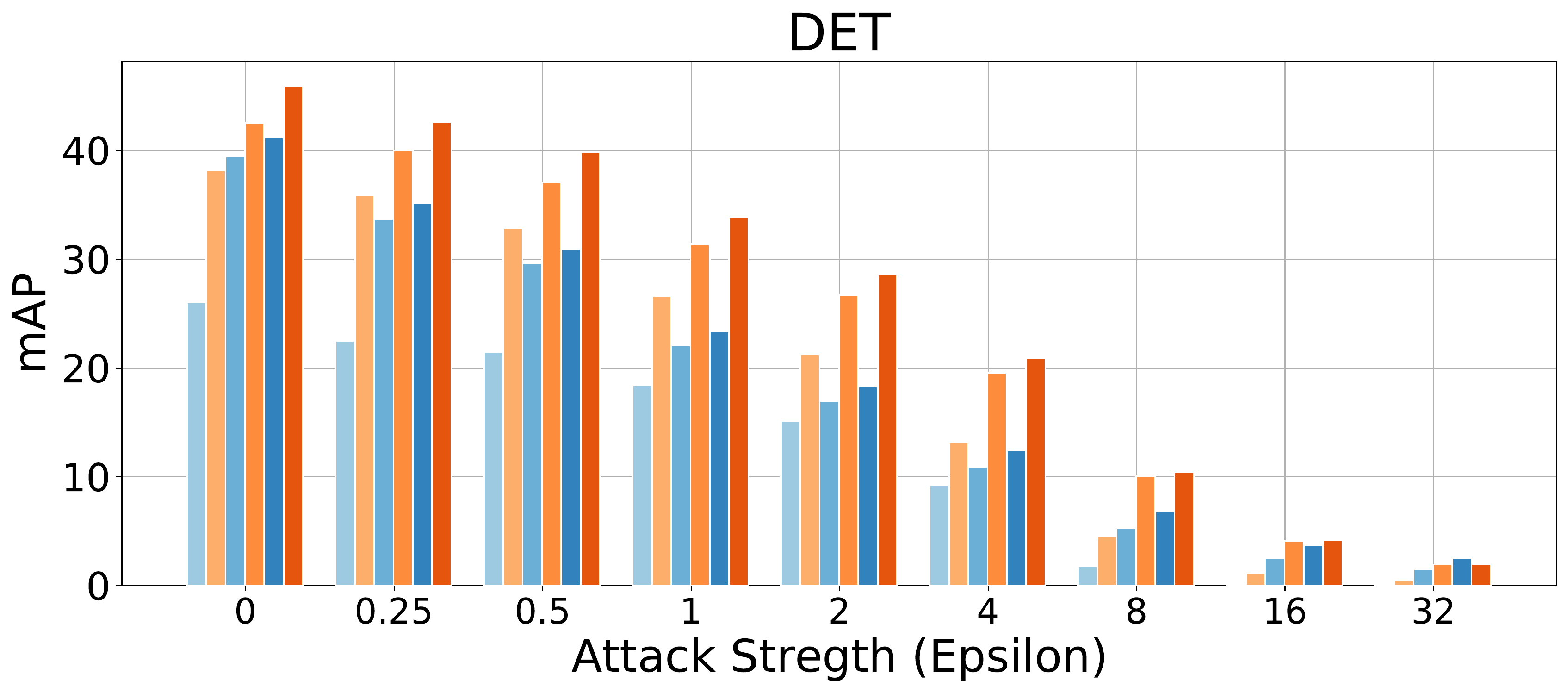}%
        \hfill
        \includegraphics[width=0.5\linewidth]{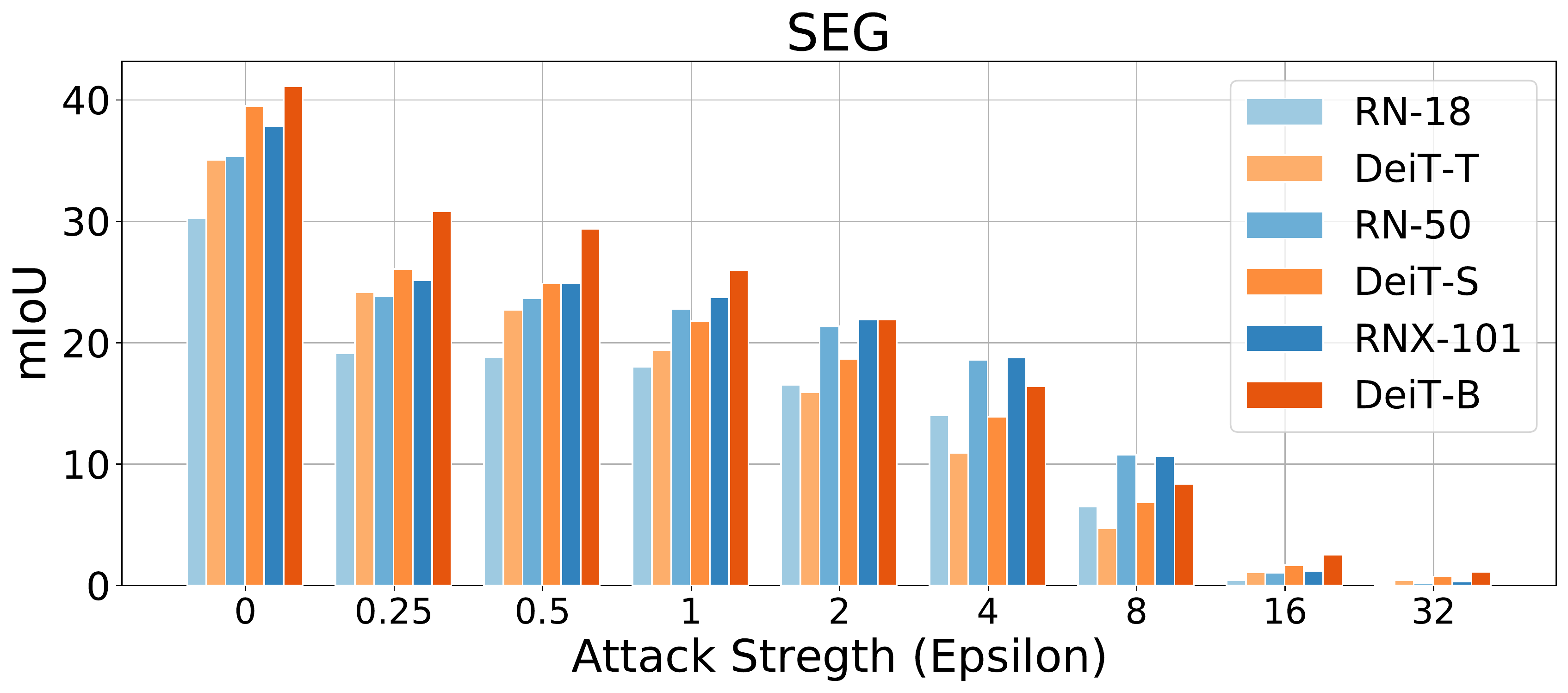}
    \end{subfigure}
    \caption{Performance comparison of VTs and their CNN counterparts under untargeted attack.}
    \label{fig:adversarial_epsilon}
\end{figure*}

\section{Robustness}
Models deployed in an ever-changing environment are exposed to natural transformations resulting from weather, lighting, or camera noise, as well as malicious transformations designed by adversaries to fool the network.
It is therefore important to evaluate the robustness of the model to natural corruptions and adversarial attacks, especially for safety-critical applications such as autonomous driving.
Thus, we evaluate the robustness of VTs and CNNs to natural corruption and adversarial attacks when used as a feature extractor in detection and segmentation.

\subsection{Natural Corruption}

\begin{figure*}[htb]
\begin{center}
   \includegraphics[width=.9\linewidth]{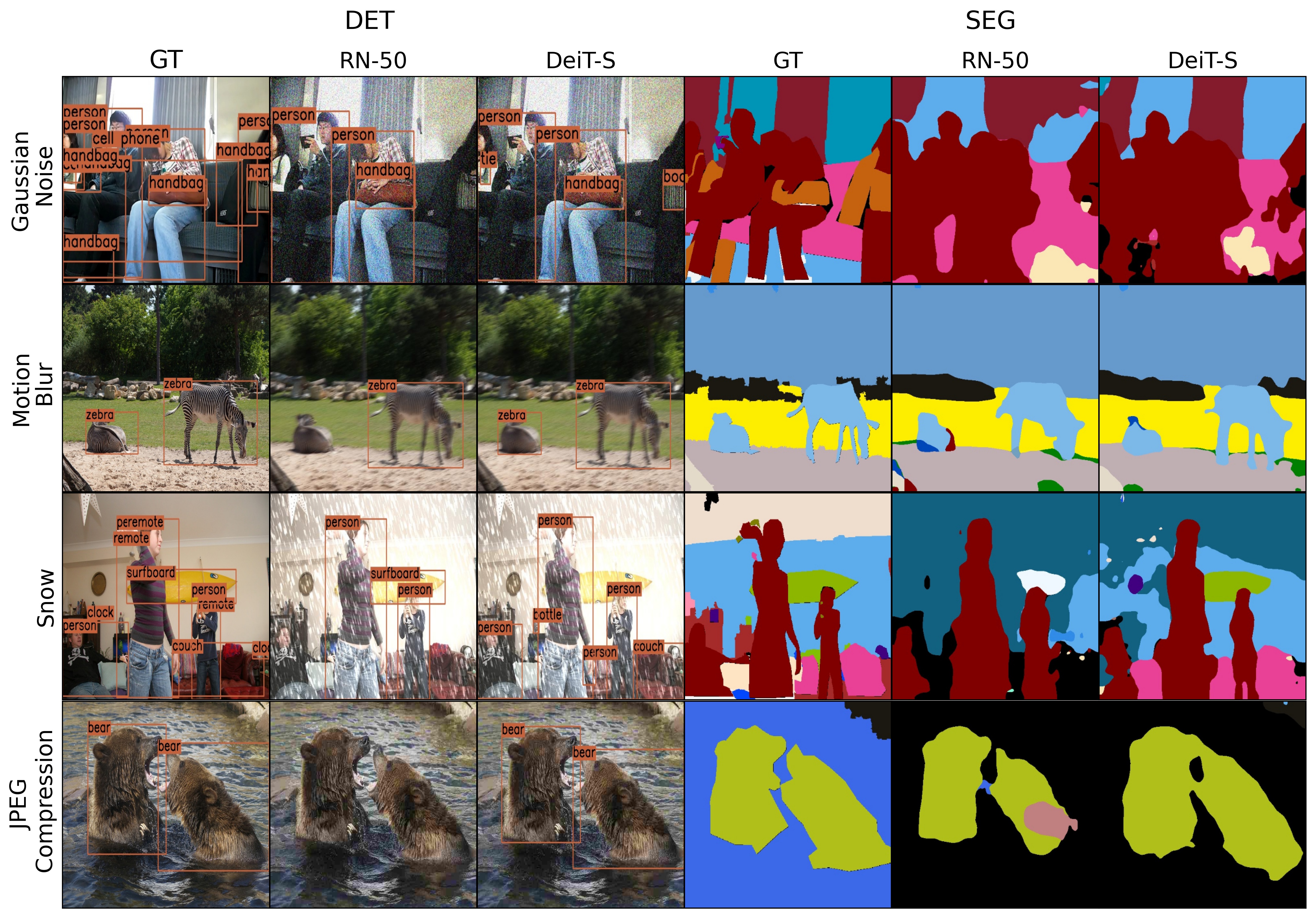}
\end{center}
   \caption{Qualitative comparison of Deit-S and RN-50 predictions on sample corrupted images. CNNs predict the wrong classes and fail to detect objects more often than VTs. For example, under `JPEG Compression' corruption, RN-50 fails to detect the objects and predicts a wrong class for part of a region in segmentation.}
\label{fig:corrupt_det}
\end{figure*}

To simulate the natural transformations in the real-world, we apply the 15 common corruptions proposed by \cite{hendrycks2019benchmarking} to the validation set of COCO and COCO-Stuff with severity 3.
We compare the performance of the VTs and CNNs on the corrupted datasets in Figure \ref{fig:corrupt} and observe that VTs are more robust than CNNs for both detection and segmentation.
This can be attributed to the global receptive field of self-attention modules in VTs that help them attend to salient and far away regions, making them less susceptible to pixel-level changes caused by corruptions.
Figure \ref{fig:corrupt_det} further provides a qualitative comparison of the model predictions for sample corrupted images.

\subsection{Adversarial Robustness}
An adversarial perturbation is an imperceptible change in the input image designed to fool the network~\cite{szegedy2013intriguing} into making a particular prediction (targeted attack) or a wrong prediction (untargeted attack). 
To generate these adversarial examples, we use the Projected Gradient Descent (PGD) attack \cite{madry2017towards} on the classification loss for both detection and segmentation. 
We use a step-size $1$ for $min(\epsilon + 4, \lceil{1.25\epsilon}\rceil)$ iterations, where $\epsilon$ is the attack strength. 
We conduct the targeted attack by swapping `person' and `car' classes.

\begin{figure}[tb]
	\centering
	\includegraphics[width=1\linewidth]{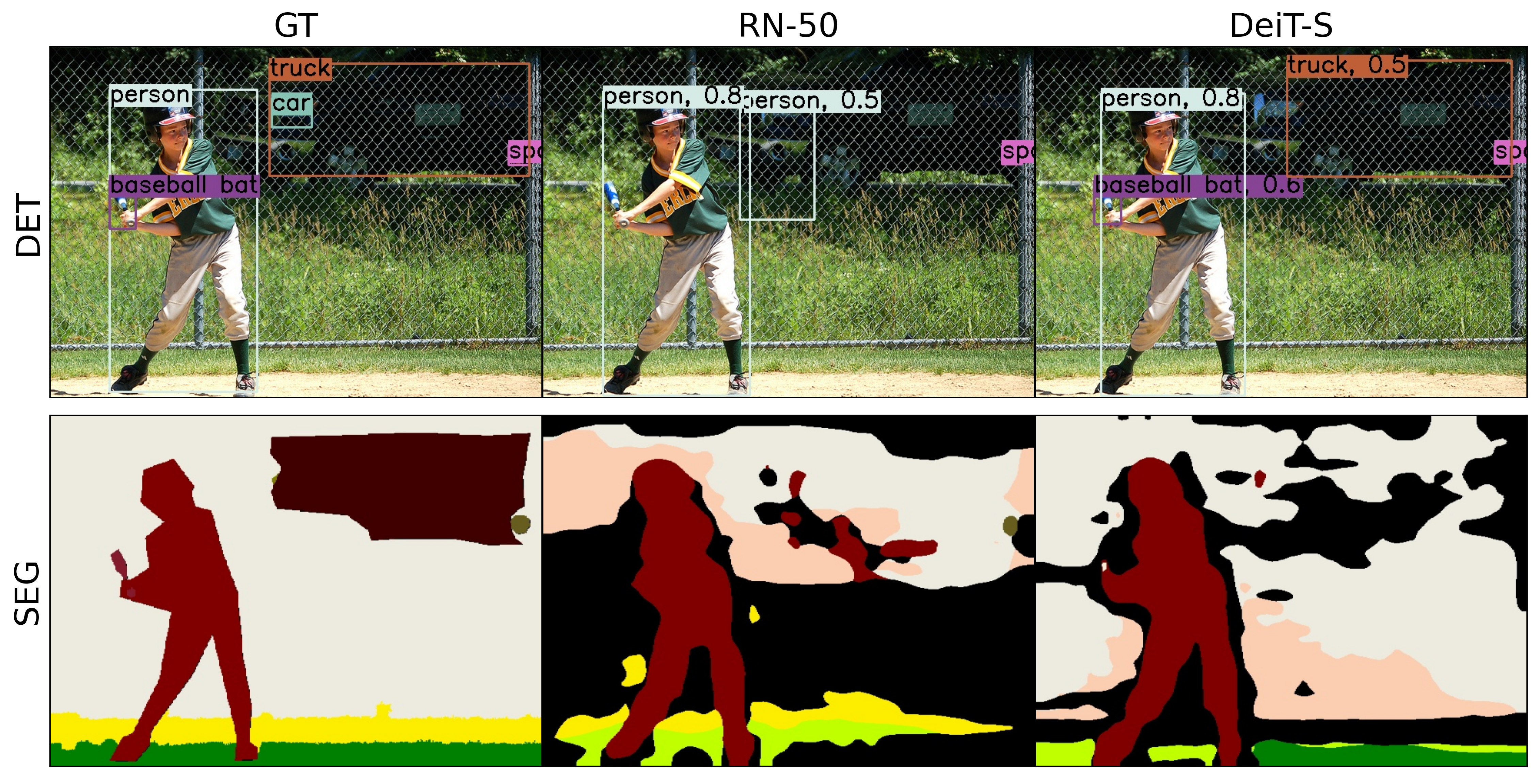}
    \caption{Qualitative comparison of DeiT-S and RN-50 predictions under targeted attack with `person' and `car' classes swapped. 
    RN-50 predicts `person' class for `car' region in detection and segmentation.
    }
    \label{fig:swap_plot}
\end{figure}

Figure \ref{fig:adversarial_epsilon} shows the performance of the models under untargetted attack at varying strengths. 
In detection, VTs are more robust to adversarial attacks compared to CNNs at all attack strengths. However, in segmentation, CNNs are more robust at higher attack strengths.
For the targeted attack, Table \ref{tab:targeted_detection} shows that VTs generally retain a higher percentage of their original AP and F1 scores compared to their CNN counterparts.
We believe that the adversarial robustness of VTs, like their robustness to natural corruptions, can be explained by the global receptive field of the self-attention modules.
Moreover, the dynamic nature of the attention modules in VTs makes it harder for the adversarial attack to find successful gradient directions to fool the network~\cite{khan2021transformers}.
Figure \ref{fig:swap_plot} illustrates the predictions by DeiT-S and RN-50 backbones for a sample targeted attacked image.

\begin{table}[tb]
    \caption{Relative performance drop in percentage of VTs and their CNN counterparts under targeted attack, when 'car' and 'person' classes are swapped. 
    Best score for each metric is in bold. The absolute performances are given in the Appendix (Table \ref{tab:targeted_attack_abs}).
    }
    \label{tab:targeted_detection}
    \centering
    \resizebox{\linewidth}{!}{%
    \begin{tabular}{l|c|c|c|c|c}
    \toprule
     & \multirow{2}{*}{Backbone} & \multicolumn{2}{c|}{`Person'} & \multicolumn{2}{c}{`Car'} \\ 
     \cline{3-6}
     &  & \multicolumn{1}{c|}{AP/IoU} & \multicolumn{1}{c|}{F1} & \multicolumn{1}{c|}{AP/IoU} & \multicolumn{1}{c}{F1} \\
     \midrule
    \multirow{6}{*}{\rotatebox[origin=c]{90}{DET}} & RN-18 & 1.88 & 2.84 & \textbf{11.68} & \textbf{9.00} \\
     & DeiT-T & 1.36 & 2.62 & 17.30 & 12.45  \\
     & RN-50 &   1.86 & 3.01 &  37.59 &  12.04 \\
     & DeiT-S &  \textbf{0.89} &  \textbf{2.11} &  17.76 &  10.54 \\
     & RNX-101 &  2.43&  2.91 &  38.47 &  9.56 \\
     & DeiT-B &  1.32 &  2.36 &  20.86 &  11.74 \\
     \midrule
    \multirow{6}{*}{\rotatebox[origin=c]{90}{SEG}} & RN-18 & 20.77 & 13.22 & 45.98 & 36.89 \\
     & DeiT-T & 14.83 & 9.01 & 44.59 & 35.17  \\
     & RN-50 & 23.88 & 15.19 & 49.61 & 39.88  \\
     & DeiT-S &  \textbf{6.94} &  \textbf{4.01} &  20.54 &  14.32 \\
     & RNX-101 &  17.15 &  10.48 &  39.08 &  29.92 \\
     & DeiT-B &  7.81 &  4.53 &  \textbf{15.49} &  \textbf{10.63} \\
     \bottomrule
    \end{tabular}}
\end{table}

\section{Texture Bias}
\label{sec:texture_bias}

Models which learn global shape-related features of objects are more robust and generalizable than the ones which rely on the texture of the objects \cite{geirhos2020shortcut}.
Texture and Shape biases \cite{hermann2020origins} are used to quantify the relative extent to which the models are dependent on texture and shape cues in image classification. 
Here, we extend the texture and shape bias analyses of VTs and CNNs for detection and segmentation tasks. 

\begin{table*}[ht]
		\caption{Texture-bias of VT and CNN backbones. SC represents the texture bias based on super-category classes. Best scores are in bold.}
		\label{table:texture_detection}
		\centering
		\begin{tabular}{lccccccc}
        \toprule
        Task & SC & RN-18 & DeiT-T & RN-50 & DeiT-S & RNX-101 & DeiT-B \\
         \midrule
        \multirow{2}{*}{{DET}}  
        & - & 4.83 & 3.10 & 4.19 & 2.55 & 3.54 & \textbf{2.42} \\
        & \checkmark & 21.98 & 20.74 & 21.35 & 19.81 & 21.12 & \textbf{19.06} \\ \hline
        \multirow{2}{*}{{SEG}}
        & - & 1.64 & \textbf{1.51} & 1.78 & 1.53 & 1.87 & 1.65 \\
        & \checkmark & 8.42 & 7.05 & 8.87 & 7.39 & 8.26 & \textbf{6.83} \\
        \bottomrule
        \end{tabular}
\end{table*}

\begin{figure}
		\centering
		\includegraphics[width=1\linewidth]{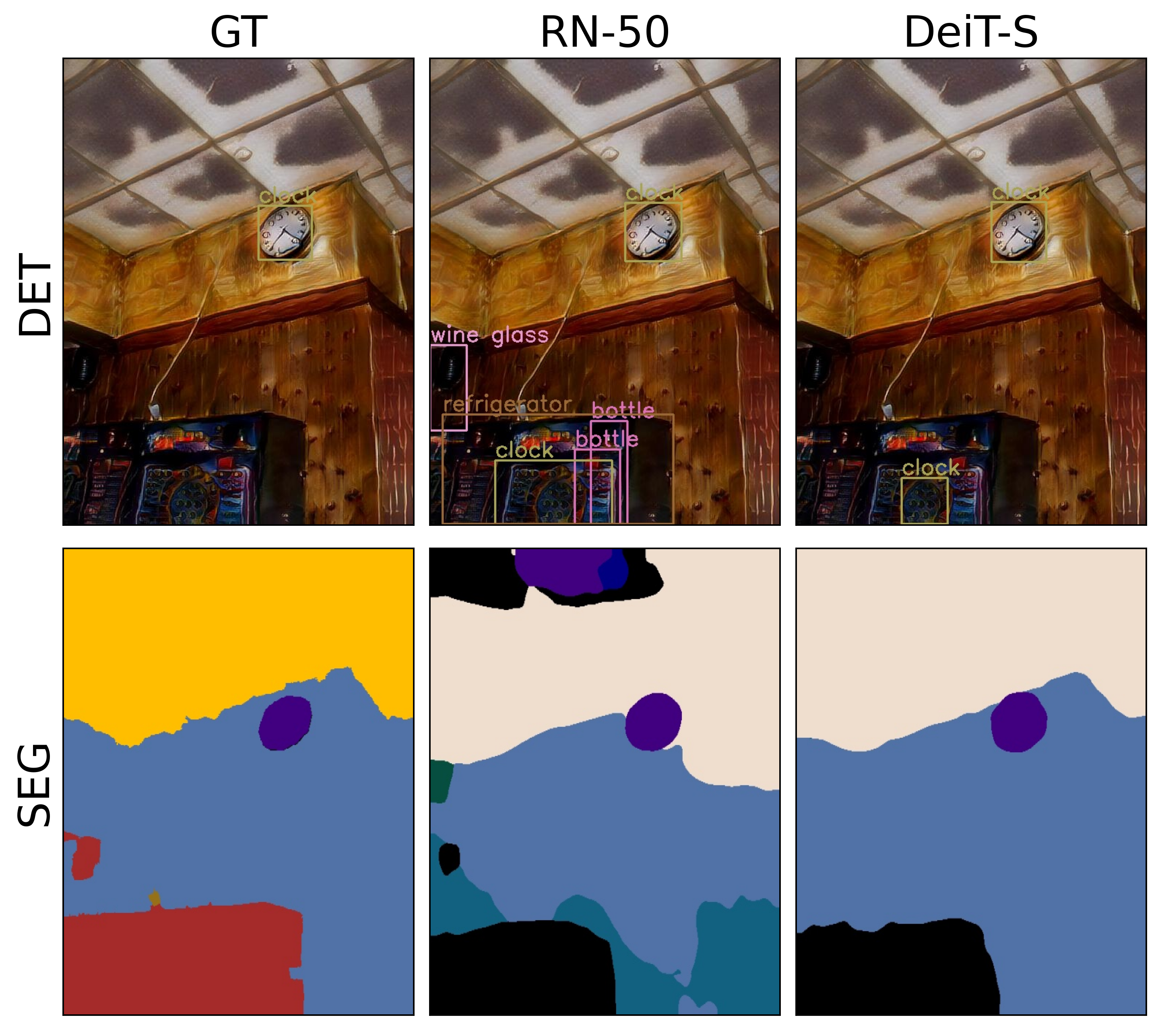}
		\caption{Qualitative comparison of DeiT-S and RN-50 predictions when the texture of `cup' class is applied on an image. RN-50 predicts `bottle' and `wine glass', which are under the same super-category as 'cup' class.}
		\label{figure:stylization}
\end{figure}

We create a texture-conflict dataset of COCO and COCO-Stuff by applying rich texture from objects (such as bear and zebra) as a style to other validation images containing multiple objects of a single class.
A model is said to predict texture in this texture-conflict dataset if it predicts the class of the applied texture. Similarly, the model is said to predict shape if it predicts the original class despite the change in texture. For $T$ texture predictions and $S$ shape predictions, the texture-bias is defined as $T / (T + S)$. From Table \ref{table:texture_detection}, we observe that, unlike in classification, the texture-bias values of the models are low for detection and segmentation. This is because while the models do not predict the shape, i.e. the intended class, they also do not predict the applied texture. 
However, from the qualitative analysis in Figure \ref{figure:stylization}, we find that the predictions and the applied-texture belong to the same COCO super-category.
Therefore, to better reflect texture-bias for detection and segmentation, we use 'Texture bias-SC', which calculates texture-bias based on super-categories.

From Table \ref{table:texture_detection}, we observe that with this metric, the models show high texture-bias, which captures their incorrect predictions. Our results indicate that VTs are less texture-biased than their CNN counterparts.
This could be explained by the global receptive field of VTs, which allows for more reliance on global shape-based cues of objects as opposed to local texture-based cues. This in turn helps Transformers to learn the "intended solution"~\cite{geirhos2020shortcut} better than CNNs, and thus generalize better to unseen data.

\section{Conclusion}

We studied different aspects of VTs and CNNs as feature extractors for object detection and semantic segmentation on challenging and real-world data. The main results and key insights derived from our experiments are as follows:

\begin{itemize}
    \item VTs outperform CNNs in in-distribution dataset while having lower inference speed, but less computational complexity.
    Hence, if the GPUs are optimized for Transformer architectures, they have the potential to become dominant in computer vision.
    \item VTs generalize better to OOD datasets. Our loss landscape analysis shows that VTs converge to flatter minima compared to CNNs, which can explain their generalizability.
    \item VTs are better calibrated than CNNs, which makes their predictions more reliable for deployment in real-world applications. 
    Moreover, we find that architecture plays the primary role in determining model calibration.
    \item Although VTs have global receptive field, their performance degrades for higher image resolutions. We believe that the interpolated positional embedding might be the reason for their performance degradation.
    \item VTs are more robust to natural corruptions and adversarial attacks compared to CNNs. We believe that this could be attributed to the global receptive field as well as the dynamic nature of self-attention.
    \item VTs are less-texture biased than CNNs, which can be attributed to their global receptive field, allowing them to focus better on global shape-based cues as opposed to local texture-based cues. 
\end{itemize}

These results and insights provide a holistic picture of the performance of both architectures, which can help the AI community make an informed choice based on the vision application.

\bibliographystyle{apalike}
{\small
\bibliography{example}}

\begin{figure*}[t]
    \centering
    \begin{subfigure}[b]{\textwidth}
        \centering
        \includegraphics[width=0.5\linewidth]{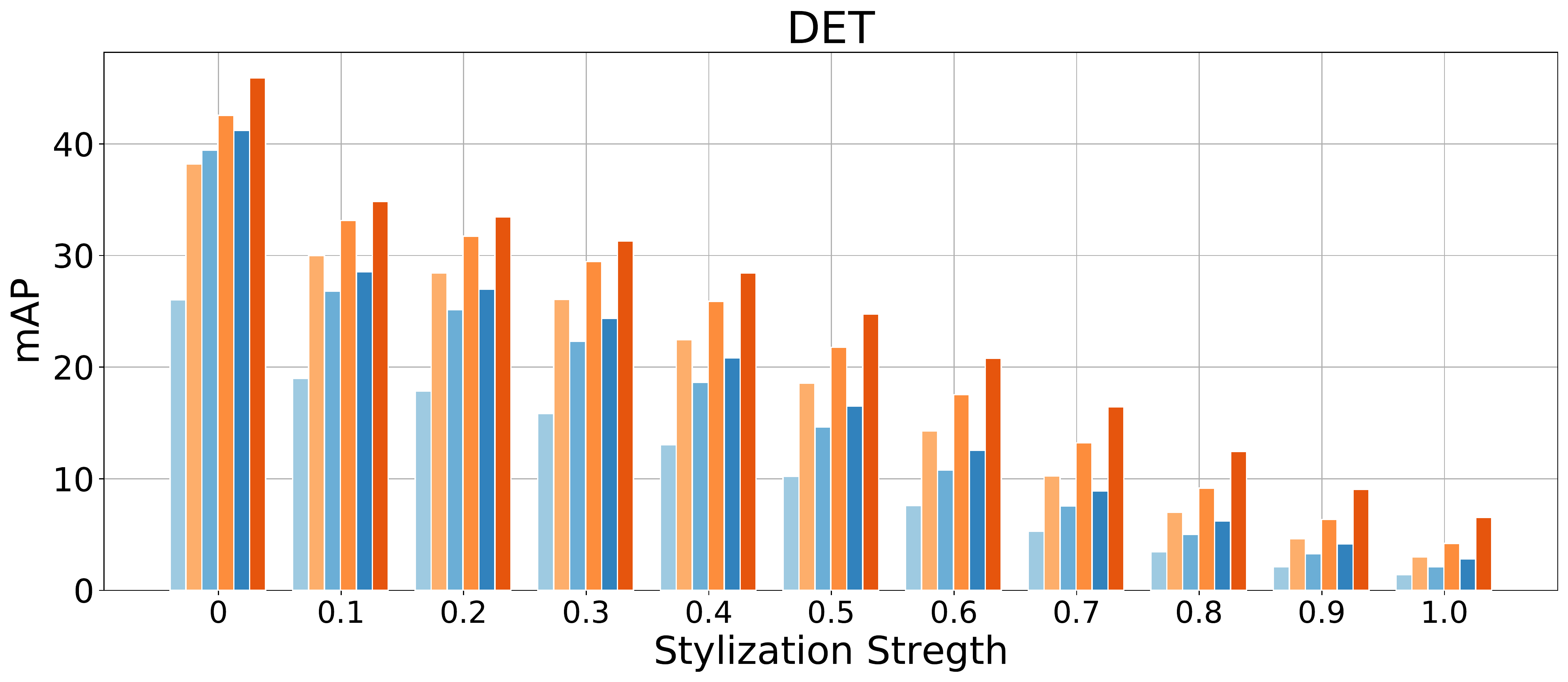}%
        \hfill
        \includegraphics[width=0.5\linewidth]{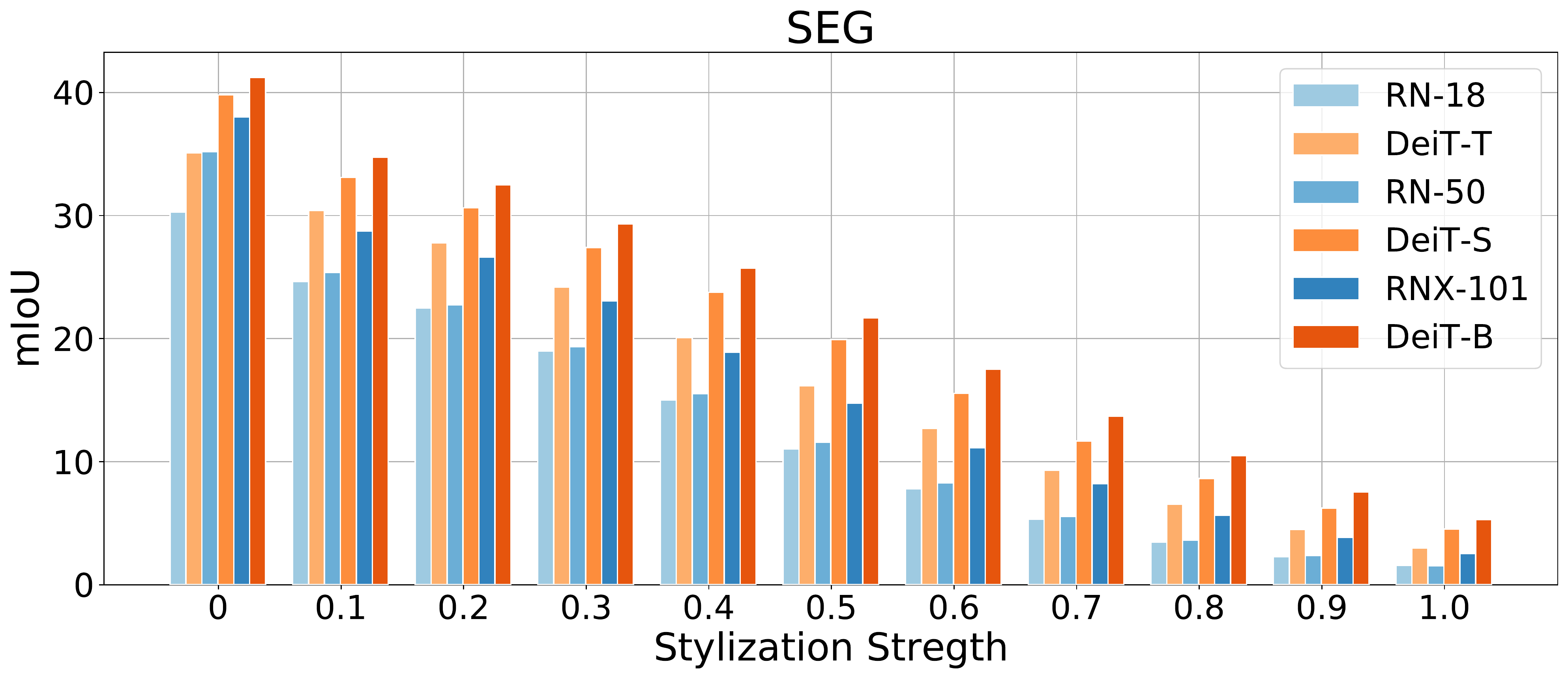}
    \end{subfigure}
    
    \caption{Performance comparison of VTs and their CNN counterparts on stylized validation sets with increasing stylization strength.}
    \label{fig:style_det}
\end{figure*}

\begin{table*}[t]
\caption{The performance of VTs and their CNN counterparts under targeted attack when ’car’ and ’person’classes are swapped}
\label{tab:targeted_attack_abs}
\resizebox{\linewidth}{!}{%
\begin{tabular}{l|c|cc|cc||cc|cc}
\toprule
 & \multicolumn{1}{l}{} & \multicolumn{4}{|c||}{`Person'} & \multicolumn{4}{c}{`Car'} \\
 \cline{3-10}
 &  & \multicolumn{2}{c|}{AP/IoU} & \multicolumn{2}{c||}{F1} & \multicolumn{2}{c|}{AP/IoU} & \multicolumn{2}{c}{F1} \\
  \cline{3-10}
 &  & B & A & B & A & B & A & B & A \\
 \midrule
\multirow{6}{*}{DET} & RN-18 & 0.7855 & 0.7707 & 1.0283 & 0.9991 & 0.4434 & 0.3916 & 0.5523 & 0.5026 \\
 & DeiT-T & 0.8705 & 0.8587 & 1.0766 & 1.0484 & 0.5833 & 0.4824  & 0.6915 & 0.6054 \\
 & RN-50 & 0.8998 & 0.8831  & 1.0910 & 1.0582  & 0.6457 & 0.4030  & 0.7600 & 0.6685 \\
 & DeiT-S & 0.8949 & 0.8869  & 1.1048 & 1.0815 & 0.6436 & 0.5293& 0.7525 & 0.6732 \\
 & RNX101 & 0.9042 & 0.8822 & 1.0920 & 1.0602 & 0.6543 & 0.4026 & 0.7623 & 0.6894\\
 & DeiT-B & 0.9090 & 0.8970 & 1.1238 & 1.0973 & 0.6798 & 0.5380 & 0.7983 & 0.7046\\
 \midrule
\multirow{6}{*}{SEG} & RN-18 & 0.7206 & 0.5709 & 0.8375 & 0.7268 & 0.4569 & 0.2468 & 0.6272 & 0.3958 \\
 & DeiT-T & 0.7578 & 0.6454 & 0.8621 & 0.7845 & 0.4817 & 0.2669 & 0.6499 & 0.4213 \\
 & RN-50 & {0.7516} & {0.5721} & {0.8581} & {0.7277} & {0.4838} & {0.2438} & {0.6521} & {0.3920} \\
 & DeiT-S & {0.7785} & {0.7245} & {0.8753} & {0.8402} & {0.5457} & {0.4336} & {0.7060} & {0.6049} \\
 & RNX101 & {0.7705} & {0.6383} & {0.8703} & {0.7791} & {0.5025} & {0.3061} & {0.6688} & {0.4687} \\
 & DeiT-B & {0.7899} & {0.7282} & {0.8826} & {0.8426} & {0.5395} & {0.4559} & {0.7008} & {0.6263} \\
 \bottomrule
\end{tabular}}
\end{table*}

\section*{\uppercase{Appendix}}

\subsection{COCO Texture Stylization for Dense Prediction Tasks}
\begin{figure}[h]
    \centering
    \includegraphics[width=\linewidth]{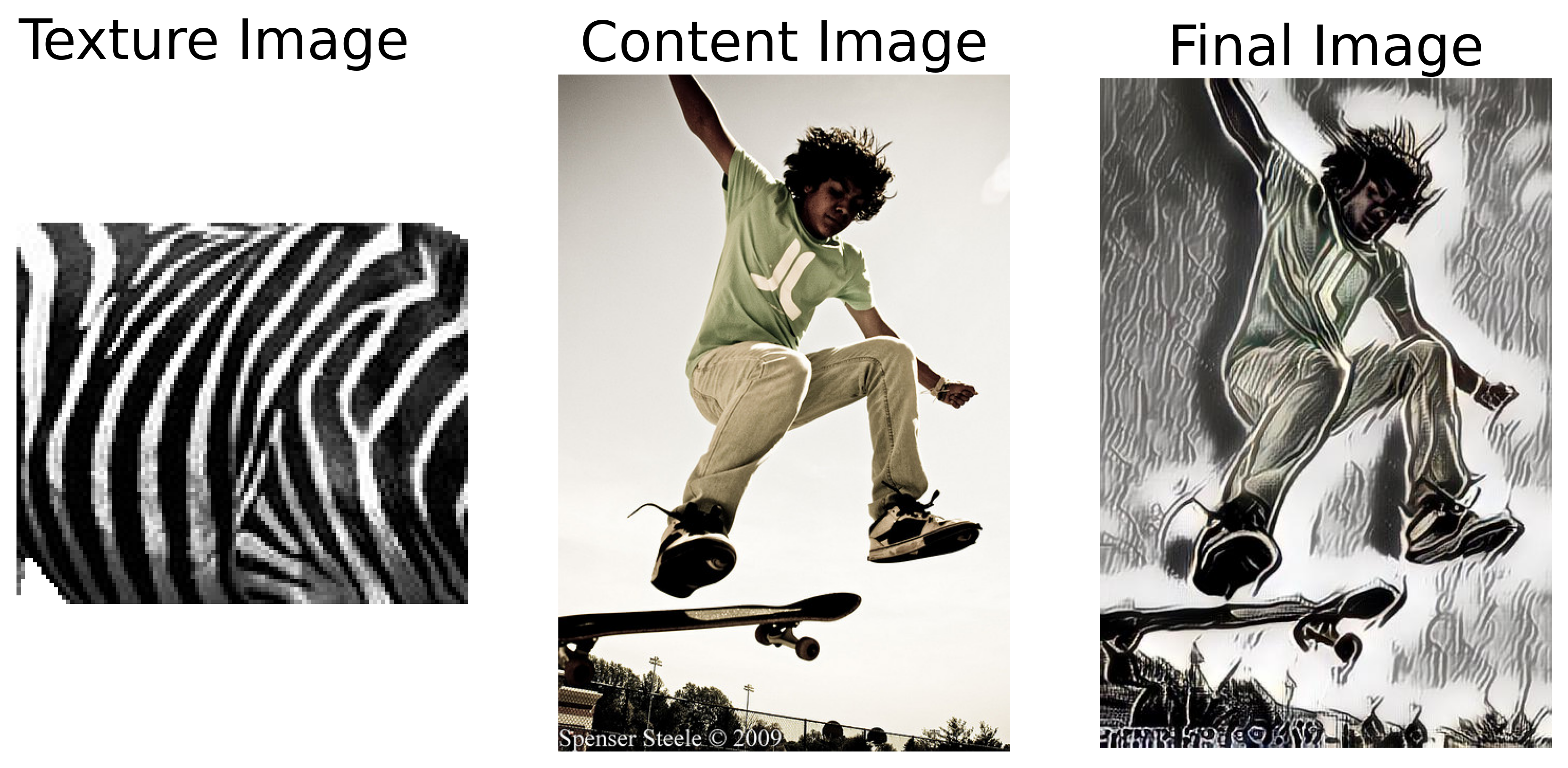}%
    \caption{Texture stylization of an image with person and skateboard with texture of zebra, at stylization strength $0.4$.}
    \label{fig:texture_approach}
\end{figure}

Here, we conduct an additional study on texture-bias of VTs and CNNs. For this, we apply a random texture from an object to every image in the COCO validation set with increasing strength of stylization using AdaIN-style~\cite{huang2017adain}\footnote{https://github.com/xunhuang1995/AdaIN-style}.
A texture is applied only if the source object is not present in the target image. Figure~\ref{fig:texture_approach} shows an example of applying a texture of a zebra on an image.
Figure~\ref{fig:style_det} demonstrates that VTs continue to rely less on texture-cues compared to CNNs at all stylization strengths. This is in line with our results in Section~\ref{sec:texture_bias}. The higher performance of VTs over CNNs with increasing stylization strength is also indicative of their higher robustness to distribution shifts.
\end{document}